\documentclass[10pt, a4paper]{article}
\usepackage{lrec}
\usepackage{multibib}
\newcites{languageresource}{Language Resources}
\usepackage{graphicx}
\usepackage{tabularx}
\usepackage{soul}

\usepackage{epstopdf}
\usepackage[latin1]{inputenc}

\usepackage{hyperref}
\usepackage{xstring}

\title{\textbf{Bianet: A Parallel News Corpus in Turkish, Kurdish and English}}

\name{Duygu Ataman}

\address{
         Fondazione Bruno Kessler\\
         Universit\`{a} degli Studi di Trento \\
         Via Sommarive 18, Trento, Italy \\
         ataman@fbk.eu\\
         }

\abstract{
We present a new open-source parallel corpus consisting of news articles collected from the Bianet magazine, an online newspaper that publishes Turkish news, often along with their translations in English and Kurdish. In this paper, we describe the collection process of the corpus and its statistical properties. We validate the benefit of using the Bianet corpus by evaluating bilingual and multilingual neural machine translation models in English-Turkish and English-Kurdish directions.  \\ \newline  }

\begin{document}

\maketitleabstract

\section{Introduction}
Machine translation (MT) is the task of translating a sequence of text to a given language. Current approaches to MT are based on statistical learning, where a probabilistic model learns to generate outputs based on previous observations of translation examples in the given language direction, often referred as \textit{parallel corpora}. By learning a statistical translation model using these corpora and using it to predict translations to future words, MT allows to automate the translation task between any pair of languages. The main advantage of statistical MT is the ability to obtain translations without requiring prior knowledge on the specific language or using any linguistic tools. On the other hand, in order to build robust and reliable translation systems it is required to use a sufficient amount of parallel corpora in the given domain of the translation that should provide observations of words and their translations in various but terminologically similar contexts. 

However, a crucial problem in statistical MT is the lack of availability of parallel corpora in many translation domains and directions. One important translation domain is news, as it has many applications in the industry, although the available parallel data in this domain is very limited in many languages. 
Turkish, for instance, is a language spoken by around 67 million people in the Republic of Turkey, although it suffers from the lack of publicly available data resources. The only parallel news corpus in the English-Turkish language direction is the South-east European Times (SETIMES) corpus \citelanguageresource{setimes}). Another language spoken in Turkey with a lack of any parallel corpora is Kurdish (Northern), with around 8 million speakers \cite{ethnologue}. Two languages are quite different by nature, Kurdish is an Indo-European language whereas Turkish belongs to the Turkic language family. However, they share many common words in their vocabularies due to a long history of social interaction between the speakers of the two languages. 

In this paper, we address this problem and present a new parallel corpus consisting of sentence-aligned news articles in Turkish, Kurdish (in the Latin script) and English. Our corpus consists of collected articles from the Bianet\footnote{Available at the website www.bianet.org} online newspaper, a website that publishes daily news on politics, law, economy and cultural events in Turkey. All articles are originally written in Turkish, while some of them are also translated into English or Kurdish (or both) by human translators, thus, they are usually available in multiple languages. We construct our corpus using online crawling tools and further process the collected sentences to check if they are correctly aligned. The retrieval process is described in Section \ref{crawling}, whereas the statistical properties of the resulting corpora are presented in Section \ref{properties}. The major part of the corpus is the English-Turkish portion, which provides additional data to translation tasks in the news domain for this language pair. Therefore, we illustrate the benefit of using the Bianet corpus in the English-Turkish language direction by building a news domain neural MT system \cite{bahdanau2014neural} and evaluate it on the data sets from an official MT evaluation campaign. Although the portions that contain Kurdish translations are not sufficient enough to build a stand-alone MT system, given the low-resource feature of Kurdish, these corpora can still be useful in applications in MT in English-Kurdish and Turkish-Kurdish language pairs, such as for building multi-lingual translation models or fine tuning generic domain models. Using this approach, we make use of the Bianet corpus for building multi-lingual neural MT models and evaluate them in English-Turkish and English-Kurdish translation directions. The findings of our experiments, given in Section \ref{experiments}, show that using our corpus for training MT systems in the news domain can aid in obtaining better translations, whereas both languages can be improved by means of multi-lingual models. The Bianet corpus is available online and can be used for any research purpose\footnote{The Bianet corpus can be downloaded from \\    \textbf{https://d-ataman.github.io/bianet}}.

\section{Collecting the Corpus}
\label{crawling}

The collection of the Bianet corpus consists of mainly three steps:

\begin{table*}[t]
\begin{center}
\begin{tabular}{|c|c|c|c|}
      \hline
       \textbf{Language} & \textbf{Number of Sentences} & \textbf{Number of Tokens} & \textbf{Vocabulary Size}\\
      \hline
       Turkish-English &35,080 & 741,080 (\textit{EN}) - 582,783 (\textit{TR}) & 61,517 (\textit{EN}) - 103,812 (\textit{TR})\\
      \hline
      English-Kurdish & 6,486 & 139,334 (\textit{EN}) - 126,350 (\textit{KU}) & 19,362 (\textit{EN}) - 21,462 (\textit{KU})\\
      \hline
     Turkish-Kurdish & 7,390 & 121,119 (\textit{TR}) - 142,668 (\textit{KU}) & 32,064 (\textit{TR}) - 23,333 (\textit{KU})\\
      \hline
\end{tabular}
\caption{Statistical Properties of the Corpus. \textit{EN:} English side. \textit{TR:} Turkish side. \textit{KU:} Kurdish side.}
\label{data}
 \end{center}
\end{table*}

\begin{itemize}
\item Crawling the Turkish news articles in the newspaper domain
\item Retrieving the document-level translations and building comparable documents in each language pair
\item Alignment of each sentence in the document-aligned corpora
\end{itemize}

In this section, we present the details of the implementation of all the steps that resulted in the Bianet corpus. 

\subsection{Crawling the news articles}
\label{crawl}

The web crawling is implemented using Scrapy\footnote{
An open source data extraction framework, available at https://scrapy.org}, an open-source library implemented in Python for extracting data from web pages. The crawling is accomplished using \textit{Spiders}, custom classes that allow to define ways to crawl pages, such as by following links or extracting portions with corresponding HTML tags. In order to crawl the news site of Bianet, we build a news Spider that continuously reads each article in Turkish by iterating over pages. From each extracted article link, the crawling continues if and only if the next article link is within the website domain and is relevant to the list of categories. The allowed categories are politics, culture, law, human rights, women, environment, society, art, sports and culture. 

For each retrieved article, the spider processes the web page source to detect if there are any links to available translations of the article. Most articles contain a link to the English and Kurdish versions in the beginning of the page, with an HTML tag that is easily discovered, as in \textit{'Click for English'}. After each article is crawled, the web pages that represent its translations are also crawled to form a group of two or three articles, each in a different language. The program extracts the raw text body from each article in the group and then saves them using the same article id. This operation is repeated for all articles in the website until all articles that fit the relevant categories are crawled. 
The overall process of crawling the website domain results in 3,214 Turkish articles which have English translations, 824 Turkish articles with Kurdish translations, and 845 English articles which also have Kurdish translations. 

\subsection{Building Comparable Documents}

The articles crawled as described in Section \ref{crawl} are later processed and combined into a collection of three portions representing each language pair. This process is quite straight-forward as our implementation of the crawling step simultaneously crawls and saves each translation in the same id as the original article. The files are cleaned and empty lines are removed before we proceed to build sentence-aligned corpora.

\subsection{Sentence Alignment}

The comparable corpora obtained by crawling the web domain and cleaning the files are later transformed into parallel corpora using a sentence aligner. In our study, we use the HunAlign sentence aligner\footnote{Available at http://mokk.bme.hu/resources/hunalign} \cite{hunalign}. Hunalign is a tool for building bilingual text at the sentence level. The program takes as input two comparable documents in different languages and then generates bilingual sentence pairs.

\section{Statistical Properties}
\label{properties}

In this section, we present the Bianet corpus which consists of 35,080 sentences, and around 1,3 million tokens. The Turkish side of the parallel corpus contains a total vocabulary of 103,812 unique words, which is a significant vocabulary contribution for a sparse language like Turkish. The English-Kurdish and Turkish-Kurdish portions are rather smaller compared to the first portion. The English-Kurdish corpus contains 6,486 sentences, whereas the Turkish-Kurdish portion contains 7,390. Further information of the statistical properties of the corpus can be found in Table \ref{data}.

\section{Experiments}
\label{experiments}

In order to illustrate the contribution of the Bianet corpus, we conduct statistical MT experiments in the English-Turkish and English-Kurdish language directions. We evaluate the benefit of using our corpus by including it in the training of models based on neural MT, the state-of-the-art method in statistical MT \cite{bahdanau2014neural}. We first evaluate the quality of the translations in a English-Turkish translation model where we show the improvement on the output accuracy with the addition of the Bianet corpus on the translation model trained on the news domain. In the second stage, since both languages are low-resource, we train multi-lingual neural MT systems based on the approach of \cite{lakewfbk}, in order to further improve the translation quality. This section presents the details of these experiments.

\subsection{Data}

\begin{table*}[h]
\begin{center}
\begin{tabular}{|c|c|c|c|c|}
      \hline
      \textbf{Data set} & \textbf{Corpus} &  \textbf{Language} &\textbf{Sentences} & \textbf{Words}\\
      \hline
      Parallel Data & SETIMES & English-Turkish & 205,706 & 5,107,219 (\textit{EN}) -  4,589,614 (\textit{TR})\\
      (Translation Model) &  Bianet & English-Turkish &35,080 & 741,080 (\textit{EN}) - 582,783 (\textit{TR})  \\
       & Bianet & English-Kurdish &6,486 & 139,334 (\textit{EN}) - 126,350 (\textit{KU}) \\
      &  Bianet & Kurdish-Turkish & 7,390 & 142,668 (\textit{KU}) - 121,119 (\textit{TR})  \\
      \hline
      Dev & WMT dev2016 & English-Turkish & 1,001 & 22,136 (\textit{EN}) - 16,954 (\textit{TR})\\
      \hline
      Test & WMT test2016 & English-Turkish & 3,000 & 66,394 (\textit{EN}) - 54,128 (\textit{TR})\\
      \hline
\end{tabular}
\caption{Data sets used in the English-Turkish Experiments. \textit{EN:} English side. \textit{TR:} Turkish side.}
\label{t2}
\end{center}
\end{table*}

\begin{table*}[h]
\begin{center}
\begin{tabular}{|c|c|c|c|c|}
      \hline
      \textbf{Data set} & \textbf{Corpus} & \textbf{Language} & \textbf{Sentences} & \textbf{Words}\\
      \hline
      Parallel Data & Ubuntu \& GNOME & English-Kurdish & 65,357 & 206,855 (\textit{EN}) - 219,279 (\textit{KU})\\
      (Translation Model) & Bianet & English-Kurdish & 6,486 & 139,334 (\textit{EN}) - 126,350 (\textit{KU}) \\
       & SETIMES \& Bianet & English-Turkish & 240,786 & 5,848,299 (\textit{EN}) -  5,172,397 (\textit{TR})\\
      &  Bianet & Turkish-Kurdish & 7,390 & 142,668 (\textit{KU}) - 121,119 (\textit{TR}) \\
       \hline
      Dev & Sampled from Bianet & English-Kurdish & 500 & 11,311 (\textit{EN}) - 5,399 (\textit{KU}) \\
      \hline
      Test & Sampled from Bianet & English-Kurdish & 500 & 11,174 (\textit{EN}) - 5,696  (\textit{KU}) \\
      \hline
\end{tabular}
\caption{Data sets used in the English-Kurdish Experiments. \textit{EN:} English side. \textit{KU:} Kurdish side.}
\label{t3}
\end{center}
\end{table*}

In English-Turkish experiments using bilingual neural MT, we build two bilingual neural MT systems in the news domain, one system only using the SETIMES corpus \citelanguageresource{setimes}, and a second system using both SETIMES and Bianet corpora. We evaluate the two models using the official news development and testing sets from WMT 2016\footnote{The Official Shared Task of MT of News, The First Conference on MT, 2016} \cite{bojar2016findings}. In the multilingual neural MT systems in the English-Turkish direction, we also use the English-Kurdish and Kurdish-Turkish portions of the Bianet corpus. Similarly, in English-Kurdish experiments, we build a generic neural MT model using a training set consisting of the only publicly available English-Kurdish parallel datasets, Ubuntu and GNOME \citelanguageresource{ubuntu}. Since there are no available official evaluation data sets, we sample the development and the testing sets from the Bianet corpus so that they reflect the news domain. We then build a multilingual neural MT model using the English-Kurdish, English-Turkish and Turkish-Kurdish portions of the Bianet corpus.
The details of the data used in the experiments can be seen in Tables \ref{t2} and \ref{t3}.

\subsection{Models}

The neural MT models used in the evaluation are based on the Nematus toolkit \cite{nematus}. They have a hidden layer and embedding dimension of 1024, a mini-batch size of 100 and a learning rate of 0.01. The dictionary size is 40,000 for both the source and target languages. For vocabulary reduction we use the subword segmentation method described in \cite{ataman}, with the default settings and a target vocabulary size of 40,000. We train the models using the Adagrad \cite{duchi2011adaptive} optimizer, and a dropout rate of 0.1 in the input and output layers and 0.2 in the embeddings and hidden layers. During training, we shuffle the data at each epoch for a total of 50 epochs and then choose the model with the highest performance on the development set for translating the test set. We use the BLEU \cite{bleu} and chrF3 \cite{chrf} automatic evaluation metrics and the Multeval ~\cite{multeval} significance test for evaluating the accuracy of the models.

In English-Turkish translation, we train two models using two different parallel corpora, one only using the SETIMES corpus, called as the \textit{Baseline Model News}, and a second model that is trained using also the Bianet corpus, referred to as \textit{Combined Model News}. This allows us to illustrate the benefit of using our corpus for Turkish translation. The multi-lingual model is referred to as \textit{Multilingual Model News}. In English-Kurdish translation, we build two models, one generic bilingual English-Kurdish model, called \textit{Bilingual Model Generic}, which uses only the English-Kurdish parallel corpora. We also build one multilingual Turkish-Kurdish-English neural MT system, \textit{Multilingual Model Generic}, which uses all available corpora. Since the already available English-Kurdish data in the IT domain (Ubuntu \& GNOME) are not sufficient to build a reliable neural MT model in the news domain, the generic models can better illustrate the performance of MT systems in this language direction.

\subsection{Results}

The translation accuracy obtained on the WMT Turkish testing sets are given in Table \ref{results1}. The model using the extended parallel training 
corpus achieves a significant improvement of \textbf{2.27 BLEU} and \textbf{0.0204 chrF3} points over the baseline model trained using only previously available SETIMES corpus. The significant improvement of 19.74\% on the given evaluation task verifies the quality of human translations in the Bianet corpus and confirms the benefit of its usage for training MT systems in the news domain. Yet, in English-Turkish translation, the best performance is achieved with the multilingual model which incorporates all portions of the Bianet corpus, which outperforms the baseline model by \textbf{2.42 BLEU} and \textbf{0.0221 chrF3} points. 

Similarly, in English-Kurdish translation, as given in Table \ref{results2}, the corpus shows promising application by allowing to generate translations in the news domain with a quality of \textbf{5.41 BLEU} and \textbf{0.2257 chrF3} points using only a generic model, which is trained on small corpora from different domains. Adding also the multi-lingual data in Turkish-Kurdish and Turkish-English from the Bianet corpus moreover increases this quality by \textbf{4.10 BLEU} and \textbf{0.0149 chrF3} points. 

\begin{table}[t]
\begin{center}
\begin{tabularx}{\columnwidth}{|c|XX|}
      \hline
      \textbf{System} & \multicolumn{2}{c|}{\textbf{Output Score}} \\
      & \textbf{BLEU} & \textbf{chrF3} \\
      \hline
      Baseline Model News & 11.50 & 0.4139\\
      \hline
      Combined Model News & 13.77 & 0.4343\\
      \hline
       \textbf{Multilingual Model News} & \textbf{13.92} & \textbf{0.4360}\\
      \hline
\end{tabularx}
\caption{English-Turkish Experiment Results. \textit{Baseline Model News:} The model built only using SETIMES corpus. \textit{Combined Model News:} The model built using SETIMES and the English-Turkish portion of Bianet corpus. \textit{Multilingual Model News:} The model built using SETIMES and all portions of the Bianet corpus. Best scores are in bold font. All improvements over the baseline are statistically significant (p-value~$<$~0.05).}
\label{results1}
\end{center}
\end{table}

\begin{table}[t]
\begin{center}
\begin{tabularx}{\columnwidth}{|c|XX|}
      \hline
      \textbf{System} & \multicolumn{2}{c|}{\textbf{Output Score}} \\
      & \textbf{BLEU} & \textbf{chrF3} \\
      \hline
      Bilingual Model Generic & 5.41 & 0.2257\\
      \hline
      \textbf{Multilingual Model Generic} & \textbf{8.51} & \textbf{0.2406}\\
      \hline
\end{tabularx}
\caption{English-Kurdish Experiment Results. \textit{Bilingual Model Generic:} The model built using corpora in English-Kurdish. \textit{Multilingual Model Generic:} The model built using English-Kurdish corpora and all languages in the Bianet corpus. Best scores are in bold font.}
\label{results2}
\end{center}
\end{table}

\section{Conclusion}
We have presented a new parallel corpus in the news domain that aims at improving the MT of Turkish and Kurdish, two very low-resource languages. Our parallel corpus is a collection of news articles retrieved from the online news magazine Bianet. We have described the process of collecting and building the corpus as well as its statistical characteristics. We have also evaluated the quality of the translations in the corpus and the advantage of using it in MT by means of a set of experiments that compare bilingual and multi-lingual neural MT models using parallel corpora with and without including the Bianet corpus. The experiment findings show that the addition of the Bianet corpus yields a significant improvement on the overall translation quality, proving that it could be useful for building MT systems in the given language pairs. Our corpus is available online for public use.


%
%
%
%
%


\section{Bibliographical References}
\label{main:ref}

\bibliographystyle{lrec}
\bibliography{xample}

\section{Language Resource References}
\label{lr:ref}
\bibliographystylelanguageresource{lrec}
\bibliographylanguageresource{xample}

\end{document}